\pdfoutput=1
\documentclass[letterpaper]{article} % DO NOT CHANGE THIS
\usepackage{aaai23}  % DO NOT CHANGE THIS
\usepackage{times}  % DO NOT CHANGE THIS
\usepackage{helvet}  % DO NOT CHANGE THIS
\usepackage{courier}  % DO NOT CHANGE THIS
\usepackage[hyphens]{url}  % DO NOT CHANGE THIS
\usepackage{graphicx} % DO NOT CHANGE THIS
\usepackage{natbib}  % DO NOT CHANGE THIS AND DO NOT ADD ANY OPTIONS TO IT
\usepackage{caption} % DO NOT CHANGE THIS AND DO NOT ADD ANY OPTIONS TO IT
\usepackage{algorithm}
\usepackage{algorithmic}

\usepackage{amsmath}
\usepackage{multirow}
\usepackage{amssymb}

\usepackage{newfloat}
\usepackage{listings}
\DeclareCaptionStyle{ruled}{labelfont=normalfont,labelsep=colon,strut=off} % DO NOT CHANGE THIS
\floatstyle{ruled}
\newfloat{listing}{tb}{lst}{}
\floatname{listing}{Listing}
\title{Learnable Blur Kernel for Single-Image Defocus Deblurring in the Wild}

\author {
    Jucai Zhai \textsuperscript{\rm 1},
    Pengcheng Zeng \textsuperscript{\rm 1},
    Chihao Ma \textsuperscript{\rm 1},
    Yong Zhao \textsuperscript{\rm 1}\thanks{corresponding author.},
    Jie Chen\textsuperscript{\rm 1,2}
}

\affiliations {
    \textsuperscript{\rm 1} Shenzhen Graduate School, Peking University \
    \textsuperscript{\rm 2} Peng Cheng Laboratory\\
    \{jucaizhai, zpceng, machihao\}@stu.pku.edu.cn, yongzhao@pkusz.edu.cn, chenj@pcl.ac.cn
}

\usepackage{bibentry}
\begin{document}

\maketitle

\begin{abstract}
Recent research showed that the dual-pixel sensor has made great progress in defocus map estimation and image defocus deblurring.
However, extracting real-time dual-pixel views is troublesome and complex in algorithm deployment.
Moreover, the deblurred image generated by the defocus deblurring network lacks high-frequency details, which is unsatisfactory in human perception. To overcome this issue, we propose a novel defocus deblurring method that uses the guidance of the defocus map to implement image deblurring.
The proposed method consists of a learnable blur kernel to estimate the defocus map, which is an unsupervised method, and a single-image defocus deblurring generative adversarial network (DefocusGAN) for the first time.
The proposed network can learn the deblurring of different regions and recover realistic details. We propose a defocus adversarial loss to guide this training process.
Competitive experimental results confirm that with a learnable blur kernel, the generated defocus map can achieve results comparable to supervised methods.
In the single-image defocus deblurring task, the proposed method achieves state-of-the-art results, especially significant improvements in perceptual quality, where PSNR reaches 25.56 dB and LPIPS reaches 0.111.

\end{abstract}

\section{Introduction}
Defocus blur occurs when a scene point outside the depth-of-field (DoF) of the lens is out-of-focus (OoF) during a camera capture \cite{abuolaim2020DPdefocus}. As shown in Figure 1, objects located at different depths have different degrees of blur. During the shooting process of the camera, the light from the scene point on the focal plane of the camera's object side is focused on the image plane, and no blur occurs. As the distance between the scene point and the focal plane of the object side of the camera gets farther, the projection of the scene point on the image plane also presents a larger circle of confusion (CoC), resulting in defocus blur. The spatial extent of the CoC can be described by a point spread function (PSF). 
\begin{figure}[t]
\centering
\includegraphics[width=1\columnwidth]{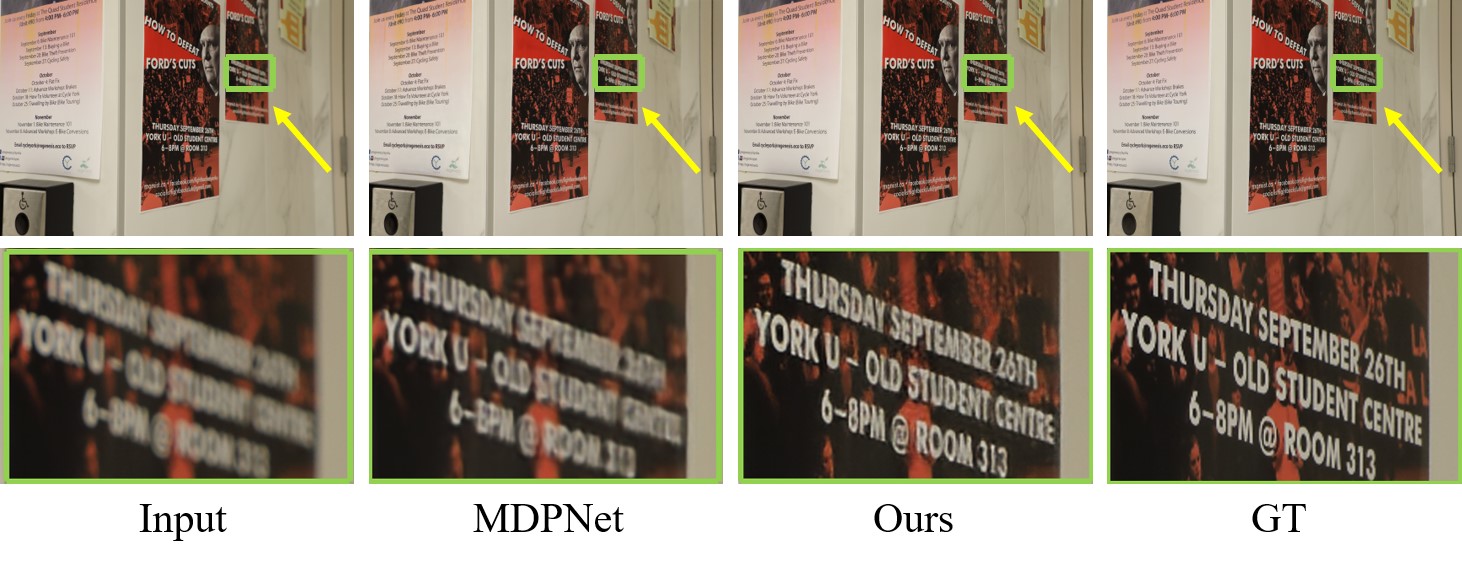} % Reduce the figure size so that it is slightly narrower than the column. Don't use precise values for figure width. This setup will avoid overfull boxes.
\caption{Qualitative comparison on the DPDD dataset. This image has obvious depth information. The first and last columns show defocused input images and their ground truth (GT). MDPNet is the best single-image defocus deblurring network. As we can see, compared to MDPNet, our generated images can handle large-scale defocus blur, recovering good structure and texture.}
\label{fig1}
\end{figure}

Most existing methods \cite{Karaali2018EBDB, Lee2021SDPdeblur} start with single-image itself to reduce defocus blur. However, according to the PSF, defocus deblurring is related to monocular depth estimation. Both obtain depth clues from an image, which is an unreliable estimation. Therefore, it is a challenge to remove the defocus blur and restore the all-in-focus (AiF) image. Recent work in \cite{abuolaim2020DPdefocus} proposed a method to remove defocus blur using the left and right views of a dual-pixel (DP) sensor as input. The idea comes from the way DP sensors work, which is similar to stereo views to provide defocus clues.
However, extracting real-time DP views is troublesome and complex in algorithm deployment\cite{Abuolaim2021RDPD}.
Existing single-image deblurring networks generate images of poor quality and cannot properly handle defocus blur due to the lack of reliable defocus clues. At the same time, there is also a lack of high-frequency (HF) information, which is unsatisfactory in human perception. It can be seen from the MDPNet \cite{Abuolaim2022MDPnet} in Figure 1. 

Based on these findings, we propose a new method to alleviate the defocus blur problem. We first generate a defocus map using DP views to obtain defocus clues. Since the current defocus map has no ground truth, we propose a learnable blur kernel (BK) to estimate the defocus map in an unsupervised way. We then propose DefocusGAN guided by defocus map. Since the defocus blur area is regular, we design a defocus adversarial loss to focus on the learning of blurred regions. Due to the use of an annealing strategy, the defocus map can be removed during inference to achieve single-image deblurring. 

The learnable blur kernel can simulate the real blur process, which simplifies the blur kernel calibration process \cite{Xin2021blurkernelmodel} and achieves better defocus map estimation. The defocus clues brought by the defocus map can guide the network to deal with the amount of blur in different regions. GAN can enrich the HF information of images, bringing more realistic structure and details. The proposal of defocus loss allows the model to concentrate on learning defocused areas.

Our main contributions are summarized as follows: 
\begin{itemize}
\item We propose a learnable blur kernel that uses DP views to estimate defocus maps via a self-supervised learning method that does not require calibration of the blur kernel. The defocus map generated by the proposed method is comparable to the current advanced supervised learning method.
\item We propose DefocusGAN for the first time, a defocus map guided multi-scale defocus deblurring GAN. Compared with previous methods, the proposed method can maintain the information of clear areas, recover the texture and details of blurred areas, and generate more realistic images.
\item The experiment results show that the proposed method is effective, with a small number of parameters, and achieves state-of-the-art performance in the single-image defocus deblurring task, where PSNR reaches 25.56 dB and LPIPS reaches 0.111.
\end{itemize}

\section{Related Works}

\subsection{Defocus deblurring}

The methods of defocus deblurring are generally divided into two categories. One class of methods is a two-stage cascade method \cite{Shi2015JNB}, which first estimates the defocus map and then deblurs the blurred image through non-blind deconvolution \cite{Fish1995deconvolution, krishnan2009deconvolution} guided by the defocus map. Another class of methods, such as DPDNet \cite{abuolaim2020DPdefocus}, MDPNet, restore the AiF image directly from the blurred image.

In the two-stage method, the difference between blurred and sharp images is used for defocus map estimation, and then deconvolution is used to restore the defocus regions. \cite{Shi2015JNB, DA2016defocusmap, Yi2016LBP, Karaali2018EBDB} used hand-crafted features to estimate defocus maps from edge differences between sharp and blurred images. \cite{Park2017DHfeatureforDE} estimated the amount of edge blur by combining deep features and handcrafted features. \cite{Lee2019DMENet} proposed a large dataset to estimate densely defocus maps. \cite{Xin2021blurkernelmodel} estimated the defocus map using a calibrated BK in an unsupervised way. \cite{Liang2021BAMBNet} used the DP views to estimate the defocus map for the first time.
However, the above methods either require handcrafted features, ground truth (GT), or the calibration of BK. Using a single blurred image does not reliably estimate the defocus map. This makes the estimation of the defocus map very difficult. Thanks to recent work, \cite{Abuolaim2021RDPD} proposed a modeling method for BK. Based on the model, we use the learning method to estimate the BK, hoping to obtain better blur characteristics, and estimate the defocus map to obtain a more effective deblurring performance.

In another class of methods, \cite{abuolaim2020DPdefocus} first introduced a large DP dataset, DPDD, and proposed DPDNet, which was the first deep learning solution to the defocus deblurring problem using the defocus clues provided by DP views. Compared with manual feature design, it achieved better performance. Although DP views were initially used in autofocus tasks \cite{Abuolaim2018Autofocus, Abuolaim2020Autofocus}, more applications have been discovered, including defocus deblurring \cite{Vo2021deblur}, depth estimation \cite{Garg2019depthestimation, Punnappurath2020DPmodel, Wu2021faceDP, Kang2021faceDP}, stereo matching \cite{Zhang2020Du2net, Pan2021DDDNet}, reflection removal \cite{Punnappurath2019reflectionremoval}, synthetic DoF \cite{Wadhwa2018SoF} and motion synthesis \cite{Abuolaim2022motionsynthesis}. Subsequently, Recurrent Neural Networks \cite{Abuolaim2021RDPD} and Multiplane Image \cite{Xin2021blurkernelmodel} were also used in DP defocus deblurring task. \cite{Liang2021BAMBNet} proposed BaMBNet, a blur-aware network, which achieved better results. Since the DP views are difficult to obtain, reasoning and deploying the network are cumbersome. Some recent works have turned to utilizing DP views to assist single-image defocus deblurring. \cite{Lee2021SDPdeblur} proposed IFAN, \cite{Son_2021_KPAC} proposed KPAC, \cite{Abuolaim2022MDPnet} proposed a multi-task defocus deblurring framework. We found that none of the above networks considered the restoration of image details. Although they get an acceptable PSNR, HF details and textures are somehow missing. The purpose of deblurring is to restore clear image details, the existing work does not seem to achieve the desired fidelity.
\begin{figure*}[t]
\centering
\includegraphics[width=0.8\textwidth]{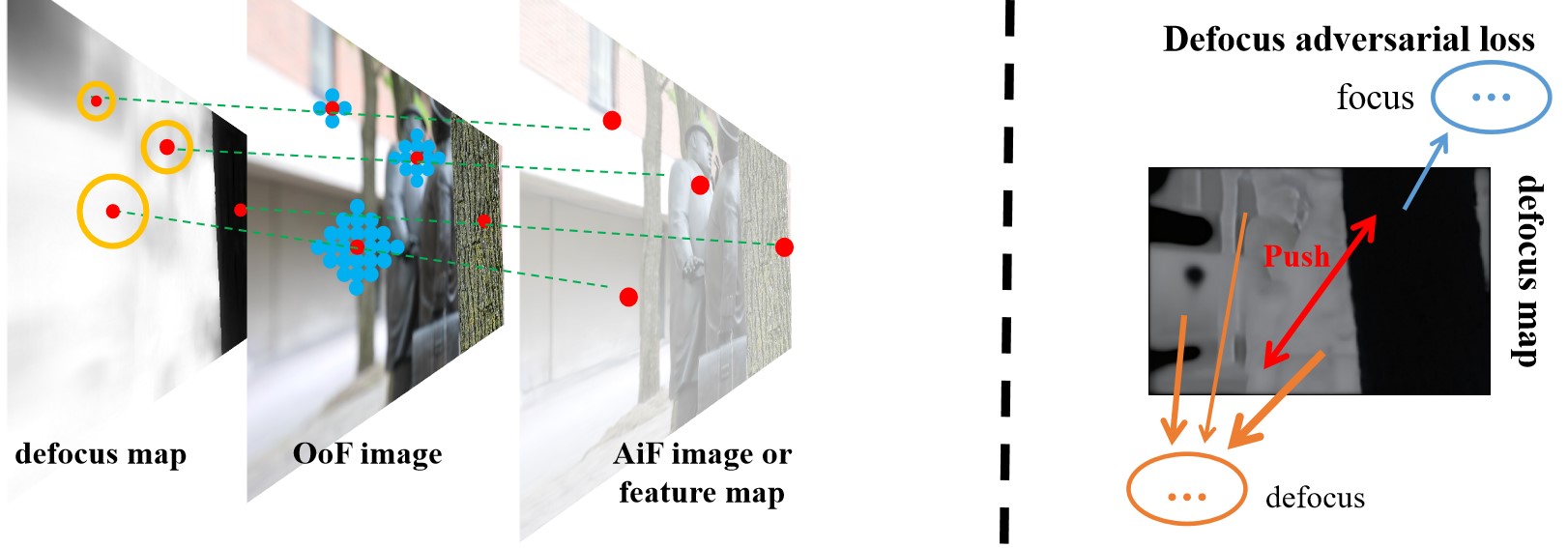} % Reduce the figure size so that it is slightly narrower than the column.
\caption{Illustration of our framework, DefocusGAN. Our framework consists of two main modules, the generator G and discriminator D. The G network takes a single input image and outputs an estimated AiF image after a multi-scale defocus deblurring block guided by a defocus map. The discriminator judges the difference with the GT. We propose a defocus adversarial loss to distinguish defocus regions within the image. The left is an illustration of G; the whiter the pixels in the defocus map, the larger the defocus parallax. The right is an illustration of defocus adversarial loss, which applies weights to different blurred areas under the guidance of the defocus map to distinguish out-of-focus and in-focus areas within the image.}
\label{fig2}
\end{figure*}

\subsection{GAN}

GAN contains two models, namely the discriminator D and the generator G, which constitutes a minimax game \cite{GOODFELLOW2014GAN, motwani2020GAN}. \cite{Ledig2017SRGAN} observes that tasks driven by L1 or L2 loss achieve high PSNR, but tend to lack HF details and are unsatisfactory in perceptual quality. GAN usually works well on details and textures. In the field of image deblurring, GAN has a very wide range of applications \cite{zhang2022Deblursurvey}, DeblurGAN \cite{kupyn2018deblurGAN} and DeblurGAN-V2 \cite{kupyn2019deblurGAN} are the most famous methods of them. While these current works were not suitable for defocus deblurring task, which achieved weak performance. Current deblurring-related GANs are based on local or global learning, whereas defocus blur is regularly regional, which has not been noticed before.

\section{Method}

\subsection{Overall Architecture}

Our model consists of two modules: one for defocus map estimation and the other for the defocus map guided multi-scale defocus deblurring (DefocusGAN) module. Figure 2 shows the illustration of our proposed framework. Defocus clues are especially important in single-image defocus deblurring task, therefore, the first step is to estimate the defocus map using DP images. Unlike BaMBNet, we propose a learnable BK and a blur reconstruction function. Then, we construct DefocusGAN. We train the network using the annealing algorithm. During the inference phase, the guide part of the defocus map can be removed and only a single-image can be used for inference.

\subsection{Learnable blur kernel for defocus map estimation}

Obtaining the spatial variation of the CoC and estimating the defocus map can effectively guide the blurred image to defocus deblur. Since the GT of the defocus map is not available, we propose an unsupervised method to estimate the defocus map. As mentioned earlier, the DP views can provide reliable defocus clues. We use the DP views for defocus parallax estimation to generate the defocus map. The value of each pixel on the defocus map represents the radius of the CoC at the current position, which is half of the defocus parallax. 
\begin{equation}
\begin{array}{*{20}{c}}
  {{I_{DPleft}}} \\ 
  {{I_{DPright}}} 
\end{array}\xrightarrow{{f({I_l},{I_r},\theta )}}{I_{DM}}\xrightarrow{{g({I_{AiF}},{I_{DM}},\varphi )}}I_{OoF}^*
\end{equation}

Let DP views $(I_{DPleft}$, $I_{DPright})$ as input, learn a network $f$ to estimate the defocus map $I_{DM}$, and then guide the AiF image to blur to obtain the blurred image $I_{OoF}^*$. This blur reconstruction process can be learned using a network $g$. $\theta$, $\varphi$ represent the learnable parameters of the network $f$ and $g$, respectively. Therefore, we propose a learnable blur kernel (BK) for blur operations. Then build a reblur geometric loss, which is the L1-Loss of the estimated blurred image $I_{OoF}^*$  and GT blurred image $I_{OoF}$:

\begin{equation}
{L_{gem}} = \left| {I_{OoF}^* - {I_{OoF}}} \right|
\end{equation}

Re-blurring requires building BK, while calibrated BK is difficult to obtain \cite{Xin2021blurkernelmodel}. Based on this intuition, we propose to learn the BK to make the BK more realistic, which does not require a calibration process. The learnable BK is constructed in this way. We need a BK that is similar to the real BK as the initial parameter and train the defocus map estimation network. After a certain stage of training, fix the parameters of the defocus map estimation network, train the defocus map estimation network and the BK alternately, and simultaneously improve the performance of the defocus map and the BK. \cite{Abuolaim2021RDPD} sampled the BK of the camera, and proposed a method to construct the BK, which is similar to the sampled BK. It is observed that the defocus BK of the camera is a high-pass filter:
\begin{equation}
    \begin{array}{l}
B(x,y) = {(1 + {(\frac{{{D_0}}}{{\sqrt {{{(x - {x_0})}^2} + {{(y - {y_0})}^2}} }})^{2n}})^{ - 1}} \circ C({x_0},{y_0})\\
{B_{init}} = G(\kappa ,\kappa ) * B
\end{array}
\end{equation}
Where $B$ is a Butterworth high-pass filter centered at $(x_0, y_0)$, $n$ is the filter order, $D_0$ controls the 3 dB cut-off frequency, $C$ is a circular function with $(x_0, y_0)$ as the center, and $ \circ $ represents the Hadamard product. Then use a gaussian kernel with a standard deviation of $ \kappa  \times \kappa $ to perform convolution smoothing. This BK is used as the initial parameter.

For the reconstructed model, pixels with different radii of the CoC, we divide the space to blur. There are:
\begin{equation}
    I_{OoF}^* = {I_{AiF}}[c(d)]*H(B(\theta );[c(d)])
\end{equation}
Where $d$ is the depth, $c$ is the estimated CoC radius, $I_{AiF}$ is the AiF image, $H$ is the blur reconstruction function $g$, and $B$ is the learnable BK. The initial parameters are $B_{init}$, and the optimal BK can be learned according to the loss function.

Since it is an unsupervised estimation, the above loss can’t well reflect the size of the CoC of pixels and the influence of noise. To alleviate this problem, we add a prior regularization term to penalize the gradient of the network output and estimate a smooth defocus map. Finally, the total loss is:
\begin{equation}
    {L_{DM}} = {L_{gem}} + \lambda \left\| {\nabla ({I_{DM}})} \right\|
\end{equation}
Where $I_{DM}$ represents the estimated defocus map, $\lambda$ is the balance factor between the geometric loss term and the regularization term. For the defocus map estimation network $f$, for simplicity, a network similar to the defocus deblurring network is used, which is introduced in the next section. 
\begin{table*}[t]
\resizebox{1\textwidth}{!}{
\begin{tabular}{l|lll|lll|lllll}
\hline
\multicolumn{1}{c|}{\multirow{2}{*}{Method}} & \multicolumn{3}{c|}{Indoor}                      & \multicolumn{3}{c|}{Outdoor}                     & \multicolumn{5}{c}{Indoor \& Outdoor}                                                              \\ \cline{2-12} 
\multicolumn{1}{c|}{}                        & PSNR↑          & SSIM↑          & LPIPS↓         & PSNR↑          & SSIM↑          & LPIPS↓         & PSNR↑          & SSIM↑          & MAE↓           & \multicolumn{1}{l|}{LPIPS↓}         & Params (M) \\ \hline
JNB                                          & 25.52          & 0.784          & 0.188          & 21.16          & 0.632          & 0.274          & 23.28          & 0.706          & 0.049          & \multicolumn{1}{l|}{0.232}          & -         \\
EBDB                                         & 25.83          & 0.790          & 0.326          & 21.21          & 0.631          & 0.407          & 23.47          & 0.708          & 0.049          & \multicolumn{1}{l|}{0.368}          & -         \\
DMENet                                       & 25.70          & 0.789          & 0.315          & 21.51          & 0.655          & 0.402          & 23.55          & 0.720          & 0.049          & \multicolumn{1}{l|}{0.360}          & 26.94     \\
DPDNet (single)                               & 26.52          & 0.828          & 0.179          & 22.08          & 0.689          & 0.229          & 24.25          & 0.757          & 0.044          & \multicolumn{1}{l|}{0.204}          & 35.25     \\
IFAN                                         & 27.80          & 0.856          & 0.131          & 22.70          & \textbf{0.719} & 0.179          & 25.18          & \textbf{0.786} & 0.041          & \multicolumn{1}{l|}{0.156}          & 10.48     \\
KPAC                                         & 28.02          & 0.852          & 0.129          & 22.64          & 0.702          & 0.190          & 25.26          & 0.774          & 0.041          & \multicolumn{1}{l|}{0.161}          & 2.06      \\
MDPNet                                       & 28.02          & 0.840          & 0.186          & 22.82          & 0.689          & 0.261          & 25.35          & 0.763          & 0.040          & \multicolumn{1}{l|}{0.225}          & 46.86     \\
Ours                                         & \textbf{28.31} & \textbf{0.857} & \textbf{0.086} & \textbf{22.94} & 0.718          & \textbf{0.135} & \textbf{25.56} & \textbf{0.786} & \textbf{0.039} & \multicolumn{1}{l|}{\textbf{0.111}} & 4.59      \\ \hline
\end{tabular}
}

\caption{Quantitative comparisons with single-image defocus deblurring methods. The best results are indicated in boldface. Results are on the DPDD dataset (the test set consists of 37 indoor and 39 outdoor scenes). }
\label{table1}
\end{table*}

\begin{figure*}[t]
\centering
\includegraphics[width=1\textwidth]{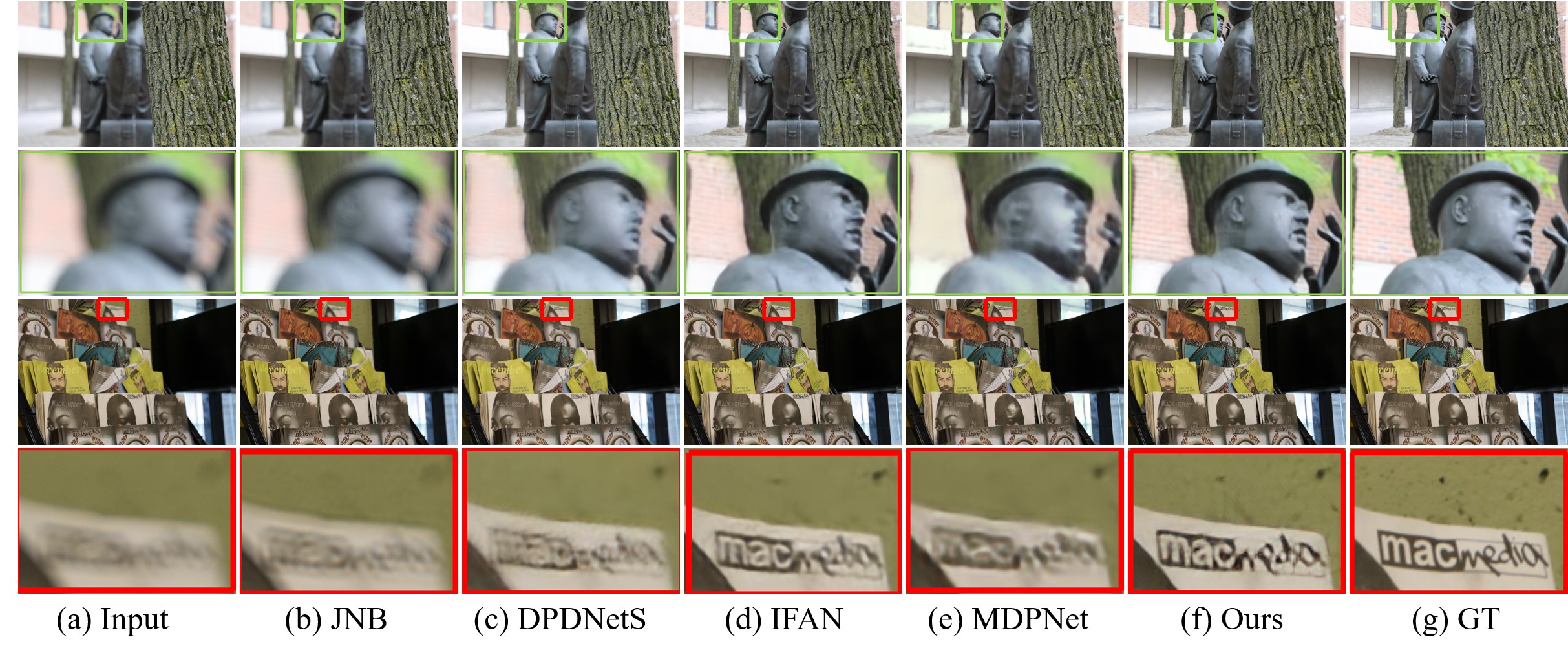} % Reduce the figure size so that it is slightly narrower than the column.
\caption{Qualitative comparison on the DPDD dataset. The first and last columns show defocused input images and their GT, respectively. In the columns, we show the deblurring results of different methods. Refer to the supplementary materials for more results.}
\label{fig4}
\end{figure*}

\subsection{Defocus deblurring GAN (DefocusGAN)}

DefocusGAN consists of a generator G and a discriminator D. We design a specialized generator network and loss function for the defocus deblurring task. Its architecture and loss will be introduced here, respectively.
\subsubsection{DefocusGAN Generator}

In the single-image defocus deblurring task, due to the lack of reliable defocus clues without using DP views, the performance on large-area blurring is poor. Inspired by this, we propose a defocus map guided multi-scale defocus deblurring network that utilizes the defocus map to provide defocus clues, constructs multi-scale layers to deal with large area blur, and uses the GAN to restore image details.

With the defocus map, a targeted defocus deblurring operation can be implemented according to the radius of the circle of confusion (CoC) corresponding to the pixel. Use the defocus mask to treat pixels with different circle sizes separately:
\begin{equation}
    \begin{array}{l}
I_{AiF}^* = \sum {(K({c_i})f({I_{OoF}},{c_j},\alpha ))} \\
\begin{array}{*{20}{c}}
{s.t.}&{K(c) = \left\{ {\begin{array}{*{20}{c}}
1&{i = j}\\
0&{i \ne j}
\end{array}} \right.}
\end{array}
\end{array}
\end{equation}
Where $K$ is a binary mask function, $f$ is the defocus deblurring function, $c$ is the radius of the CoC, and $\alpha$ are learnable parameters. Areas with the same radius of CoC can be deblurred with the same defocus deblurring function. We set $N$ deblurring branches according to the range of the radius of CoC in the defocus map. Using multiple branches to extract features, adaptive deblurring.

An illustration of our framework is shown in Figure 2. The model takes a single-image as input, passes it through a defocus map guide block (DGB), and obtains preliminary deblurred features. Then downsample to $\frac{1}{4}$ size of the input and go through the same operation. Then downsample to $\frac{1}{8}$ size of the input and repeat the operation. In this way, the cascade refinement can obtain a larger receptive field and obtain the relationship of the large-area blur range. DGB divides multiple branches according to the characteristics of the defocus map. We design 4 sets of defocus masks according to the characteristics of the defocus map and divide DGB into 4 branches. It is then multiplied by the defocus mask to obtain the features of the corresponding area. We assign a weight to the defocus mask, use the simulated annealing algorithm to reduce the weight during the training process to gradually remove the guidance of the defocus mask, and then use the prior learned by the network to continue training.

When the radius of CoC is small, the pixel where it is located can be recovered without aggregating the features of surrounding pixels. As the radius increases, the distance between the center of CoC and the current pixel increases, and a larger receptive field is required to aggregate the features of the surrounding pixels to achieve the effect of deblurring. For branches with a small radius of CoC, it is only necessary to pay attention to the pixel itself and surrounding features, and a fully convolutional network can meet the needs. For branches with a large radius, it is necessary to aggregate blurred information over a larger receptive field. Using the U-Net-like as the backbone of these branches, we replace the convolutional layer with the Residual Channel Attention Module (RCAB), further improving the ability to obtain global information.
Finally, DGB is a 4-branch network where each branch is a Unet-like structure with 8 convolutional layers. The convolutional layers of 4 branches are as follows: 1 fully convolutional layer, 1 RCAB, 2 RCAB, and 3 RCAB, respectively.
For $\frac{1}{4}$ and $\frac{1}{8}$ scale features, they have a larger receptive field after downsampling, so we appropriately reduce the parameters in DGB.

\subsubsection{DefocusGAN Discriminator}
The discriminator D is used to judge the gap between the input image and the real image, we take the output of G as the input of D and use the D like patchGAN \cite{isola2017patchgan}, which uses a 3-layer fully convolutional network to map the input image and GT to $N \times N$ matrix that compares the gap between the input image and GT. 

\subsubsection{Overall loss function}
The overall loss function used for training, especially the proposed defocus adversarial loss, has been investigated in this section.

\textbf{Defocus adversarial loss.} 
For the defocus deblurring task, we design a defocus adversarial loss that focuses on defocus regions. We reweigh the discriminator response for the first time on defocus deblurring. Due to the regularity of the defocus distribution, based on the defocus clues provided by the defocus map, we can easily get the blurred area. According to the radius of the CoC in the defocused areas, we assign different weights to the image features output by the discriminator. The larger the radius of the CoC in the defocused areas, the greater the defocus weight of the corresponding area. With this loss, the D can increase the ability to distinguish different blurred areas in the image and strengthen the learning ability of G for the blurred areas.
\begin{equation}
    {L_{defocusadv}} = \frac{1}{n}\sum\limits_{n = 1}^N { - \varphi ({I_{DM}}){D_{\theta D}}({G_{\theta G}}({I_{OoF}}))} 
\end{equation}
Where $I_{DM}$ represents the defocused image, $I_{OoF}$ represents the input defocused image, ${G_{\theta G}}$ represents the G network, and ${D_{\theta D}}$ represents the D network, $\varphi ({I_{DM}})$ represents the operation of assigning weights according to the defocus map, and ${L_{defocusadv}}$ refers to calculating the loss after assigning defocus weights to the AiF image output by discriminator D.

Similar to WGAN-GP \cite{Gulrajani2017WGAN-GP}, we add a gradient regularization term. While stabilizing GAN training, push the generative distribution towards a more realistic distribution. Instead of indiscriminately learning pictures, we focus on learning defocus regions. Similar to DeblutGANV2 \cite{kupyn2019deblurGAN}, we use L1-Loss as the content loss $L_c$. Compared to previous methods for defocus deblurring, we use a perceptual loss \cite{JOHNSON2016LP} $L_p$ to update the model, which computes the Euclidean loss on the VGG19 \cite{Simonyan2015VeryDC} $conv3{\_}3$ features maps.

We use the three losses above weighted to get ${L_G}$ to train the model, where $\alpha$ and $\beta$ are hyperparameters to balance different types of loss.
\begin{equation}
{L_G} = {L_c} + \alpha  \times {L_p} + \beta  \times {L_{defocusadv}}
\end{equation}
\section{Experiments}

\subsection{Datasets}
We use the dataset DPDD provided by \cite{abuolaim2020DPdefocus} for training and testing. This dataset has 500 sets of images, and each set of images includes a defocus blurred image, a pair of DP views, and an all-in-focus (AiF) image with a resolution of 1680 $\times$ 1120. Here, like most methods \cite{Abuolaim2021NTIRE}, flowing the settings, 500 groups have been divided into 350, 74, and 76 groups according to the training set, validation set, and test set. We also use the CUHK dataset \cite{Shi2015JNB} and the Google PixelDP dataset \cite{abuolaim2020DPdefocus} to verify the generalization of the network. Dataset conditions and training settings are described in supplementary materials.

\begin{table}[t]
\centering
\resizebox{0.90\columnwidth}{!}{
\begin{tabular}{l|llll}
\hline
Method           & PSNR↑          & SSIM↑          & MAE↓           & LPIPS↓         \\ \hline
DPDNET (DP views) & 25.13          & \textbf{0.786} & 0.041          & 0.223          \\
RDPDNet          & 25.39          & 0.772          & 0.040          & 0.179          \\
Ours             & \textbf{25.56} & \textbf{0.786} & \textbf{0.039} & \textbf{0.111} \\ \hline
\end{tabular}
}
\caption{Quantitative comparisons with some methods using DP views as input. Results are on the DPDD dataset.}
\label{table2}
\end{table}
\begin{figure}[t]
\centering
\includegraphics[width=1\columnwidth]{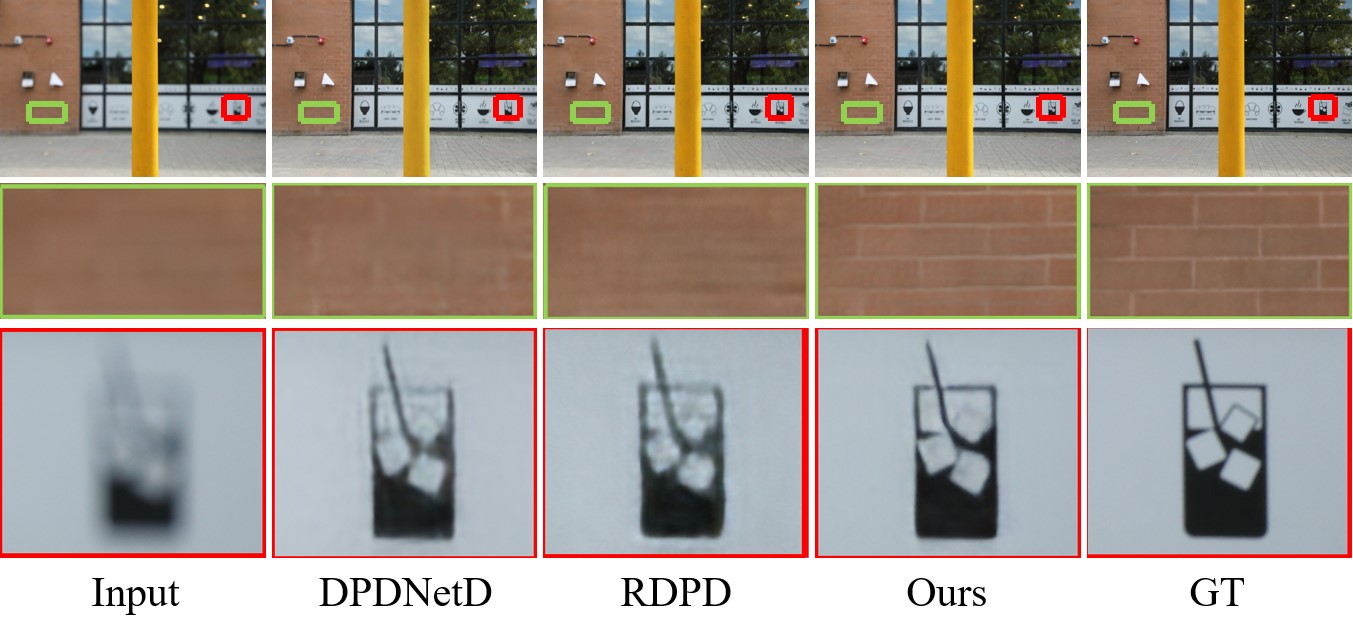} % Reduce the figure size so that it is slightly narrower than the column. Don't use precise values for figure width. This setup will avoid overfull boxes.
\caption{Qualitative comparison with methods using DP views as input on the DPDD dataset. Refer to the supplementary materials for more results.}
\label{fig5}
\end{figure}

\subsection{Implementation details}
We first train the defocus map estimation network, taking the DP views as input, to estimate the defocus map. The hyperparameter $\lambda$  of the loss function is set to $10^{-5}$, and the learning rate is set to $2 \times 10^{-5}$. First, the blur reconstruction network is fixed, and the defocus map estimation network is trained for 10 epochs. Then fix the parameters of the defocus map estimation network, and then train the blur reconstruction network. Use the method of alternating training, alternating every 5 epochs until 30 epochs.
Referring to \cite{Liang2021BAMBNet}, we set the upper limit of the radius of the CoC to 25 pixels. After the network is trained, defocus maps are obtained, which can be directly used for deblurring tasks by deconvolution or can be used as defocus clues to guide the training of the defocus deblurring network. 

Then, we train the defocus deblurring network. Here, the $512 \times 512$ single-image and the defocus map are fed into the network. The number of iterations of the simulated annealing algorithm is set to $2 \times 10^{4}$, the hyperparameters $\alpha$, $\beta $ are set to 0.012 and 0.002, respectively. The initial learning rate is set to $2 \times 10^{-4}$, which decreases by half every 30 epochs. After about 15 epochs, the model no longer relies on the guidance of the defocus map, and it gradually converges after 90 epochs. When inferring, we can discard the defocus map and use only a single-image as input to complete the defocus deblurring operation.

The batch size of both networks is set to 4 and optimized using the Adam optimizer, where b1=0.9, b2=0.999. We implemented the method using Pytorch and trained it on an NVIDIA RTX 3090 GPU.

\begin{figure*}[t]
\centering
\includegraphics[width=1\textwidth]{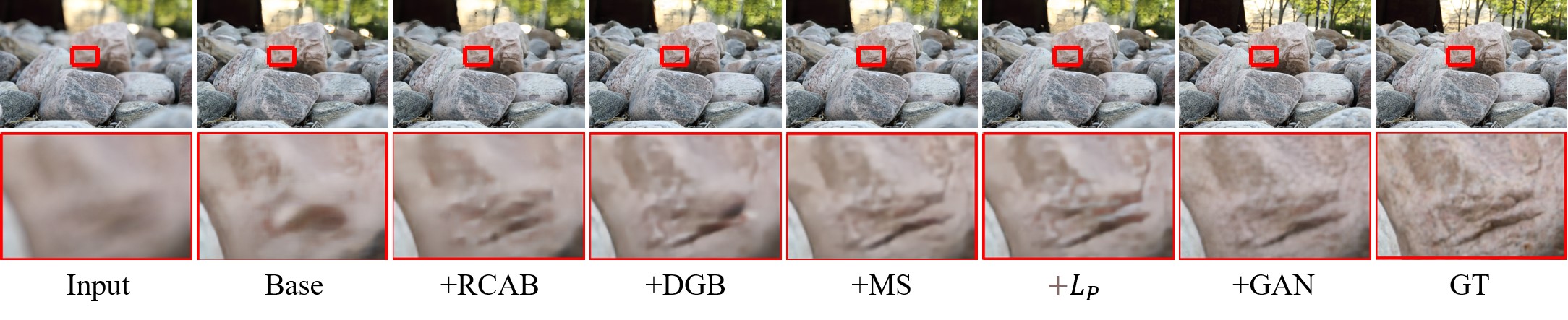} % Reduce the figure size so that it is slightly narrower than the column.
\caption{Qualitative results of an ablation study on the DPDD dataset.}
\label{fig6}
\end{figure*}

\subsection{Performance evaluation}
Like many works, to evaluate the performance of defocus deblurring, we use the test set provided by \cite{abuolaim2020DPdefocus} for testing. We compare the results with recent single-image defocus deblurring works. JNB \cite{Shi2015JNB}, EBDB \cite{Karaali2018EBDB} and DMENet \cite{Lee2019DMENet} are methods based on defocus maps. After they estimate the defocus map, they use non-blind deconvolution to defocus deblurring. DPDNet (single) \cite{abuolaim2020DPdefocus}, IFAN \cite{Lee2021SDPdeblur}, KPAC \cite{Son_2021_KPAC}, and MDPNet \cite{Abuolaim2022MDPnet} are direct estimation methods that can directly restore AiF images.
\begin{table}[t]
\centering
\resizebox{0.8\columnwidth}{!}{
\begin{tabular}{l|llll}
\hline
Method  & PSNR↑ & SSIM↑ & MAE↓  & LPIPS↓ \\ \hline
BaMBNet & 22.56 & 0.687 & 0.054 & \textbf{0.319}  \\
DMENet  & 23.55 & \textbf{0.720} & \textbf{0.049} & 0.360  \\
Ours    & \textbf{23.56} & 0.711 & \textbf{0.049} & 0.361  \\ \hline
\end{tabular}
}
\caption{Quantitative comparisons with the defocus map used for recovering AiF images on the DPDD dataset. Results are on the DPDD dataset.}
\label{table3}
\end{table}
\begin{table}[t]
\centering
\resizebox{1\columnwidth}{!}{
\begin{tabular}{llllll|lll}
\hline
\multicolumn{6}{c|}{Method}       & \multicolumn{3}{c}{Metrics} \\ \hline
Base & RCAB & DGB & MS & $Lp$ & GAN & PSNR↑   & SSIM↑   & LPIPS↓  \\ \hline
\checkmark    &      &     &    &     &    & 24.73   & 0.762   & 0.199   \\
\checkmark   & \checkmark    &     &    &     &    & 25.10   & 0.771   & 0.196   \\
\checkmark    & \checkmark    & \checkmark  &    &     &    & 25.30   & 0.782   & 0.178   \\
\checkmark   & \checkmark    & \checkmark   & \checkmark  &     &    & 25.47   & \textbf{0.786}   & 0.172   \\
\checkmark    & \checkmark    & \checkmark   & \checkmark  & \checkmark   &    & 25.44   & 0.783   & 0.115   \\
\checkmark    & \checkmark    & $\checkmark$   & \checkmark  & \checkmark   & \checkmark & \textbf{25.56}   & \textbf{0.786}   & \textbf{0.111}   \\ \hline
\end{tabular}
}
\caption{Quantitative results of the ablation experiments on the DPDD dataset.}
\label{table4}
\end{table}

For the above methods, we use the code and weights provided by the authors for testing (IFAN uses data augmentation. So we remove this and retrain according to the code and training method provided by the authors). For JNB, EBDB, and DMENet, following the advice of \cite{abuolaim2020DPdefocus}, we use the deconvolution method \cite{Fish1995deconvolution, krishnan2009deconvolution} to recover the AiF image using the estimated defocus map. We also evaluate the number of network parameters in the inference stage to characterize the size of the model.

We use the commonly used metrics PSNR, SSIM, MAE, and LPIPS for defocus deblurring to evaluate the quality of the images. Table 1 shows the quantitative results of our method and other methods. Our method shows higher quality, outperforms all current methods with few model parameters, and restores image details to a great extent, improving the realism of images. Figure 3 shows a qualitative comparison. Traditional methods based on defocus maps and deconvolution have large blur areas. The performance of MDPNet, KPAC, and IFAN is greatly improved compared with the previous results, but often produces unnatural textures such as artifacts. For example, the texture of red walls and bronze figures. In particular, compared with this method, our method can better handle the texture of the image and recover the contours of objects that conform to human subjective perception, such as text in magazines. From Figure 3, we can see that our method can better recover large-area blur, image details, and texture. More qualitative results in supplementary materials.

For completeness, we also report some methods that take DP views as input. Table 2 shows this comparison. It can be seen that, compared with DPDNet (DP views) and RDPDNet \cite{Abuolaim2021RDPD}, our method has a good performance. As shown in Figure 4, compared to these methods, we perform better, recovering images with texture details and human perception. Models are smaller and more functional. In the inference stage, only a single-image is required, but DP-based methods require access to 2 DP views. Because of the difficulty of obtaining DP views, our method has great advantages.

Since defocus maps have many practical applications \cite{Lee2019DMENet}, we compare our method with current deep learning-based methods for recovering defocus maps. We use a non-blind deconvolution method to recover AiF images with defocus maps on the DPDD dataset. It can be seen in Table 3 that, compared with BAMBNet, which is also based on the unsupervised method to recover defocus maps, we achieve great improvement. Compared with the supervised learning-based method DMENet, we achieve competitive results.
\begin{table}[t]
\centering
\resizebox{0.8\columnwidth}{!}{
\begin{tabular}{l|lll}
\hline
\multicolumn{1}{c|}{blur kernel function} & PSNR↑ & SSIM↑ & MAE↓  \\ \hline
Gaussian                                  & 22.56 & 0.687 & 0.054 \\
Butterworth                               & 23.31 & 0.702 & \textbf{0.049} \\
Ours                                      & \textbf{23.56} & \textbf{0.711} & \textbf{0.049} \\ \hline
\end{tabular}
}
\caption{Ablation study to demonstrate the effectiveness of our learnable blur kernel. Results are on the DPDD dataset.}
\label{table5}
\end{table}
\begin{figure}[t]
\centering
\includegraphics[width=0.7\columnwidth]{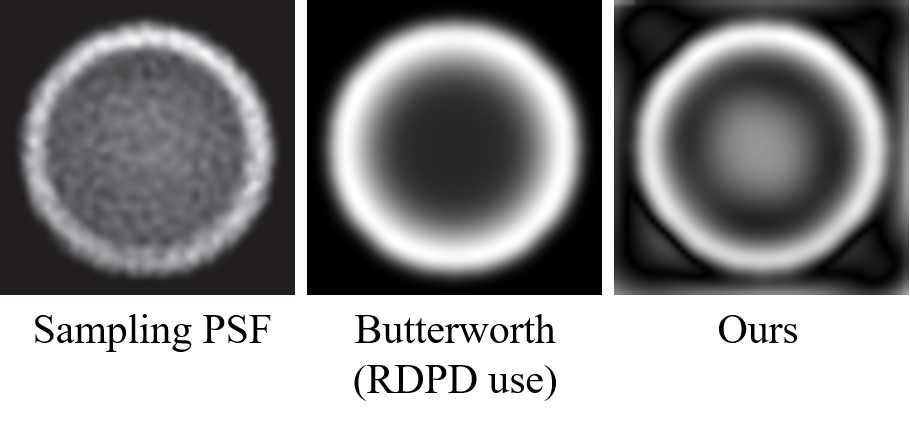} % Reduce the figure size so that it is slightly narrower than the column. Don't use precise values for figure width. This setup will avoid overfull boxes.
\caption{Qualitative comparison between different blur kernels.}.
\label{fig7}
\end{figure}

\subsection{Ablation studies}

\textbf{Effects of each module.} 
To demonstrate the effectiveness of each part of the module, we conduct ablation experiments in which all models are trained under the same conditions (e.g., optimizer, learning rate, random seed, etc). Specifically, we take a single-scale network as the baseline model. The components of the network have a defocus map guide (DG) part, RCAB, multi-scale (MS) module, perceptual loss ($L_p$), and GAN. Used components are recovered gradually from the baseline model.
\begin{table}[t]
\resizebox{1\columnwidth}{!}{
\begin{tabular}{ll|llll}
\hline
\multicolumn{2}{l|}{Method}  & \multicolumn{4}{l}{Metrics}    \\ \hline
generator & discriminator    & PSNR↑ & SSIM↑ & MAE↓  & LPIPS↓ \\ \hline
FPN       & w original loss  & 24.23 & 0.746 & 0.045 & 0.150  \\
Ours      & w original loss  & 25.31 & 0.780 & 0.040 & 0.149  \\
Ours      & w doubleGAN loss & 24.92 & 0.763 & 0.041 & \textbf{0.132}      \\
Ours      & w defocus loss   & \textbf{25.42} & \textbf{0.784} & \textbf{0.039} & 0.160  \\ \hline
\end{tabular}
}
\caption{Ablation study to demonstrate the effectiveness of GAN. Results are on the DPDD dataset.}
\label{table6}
\end{table}
\begin{figure}[t]
\centering
\includegraphics[width=1\columnwidth]{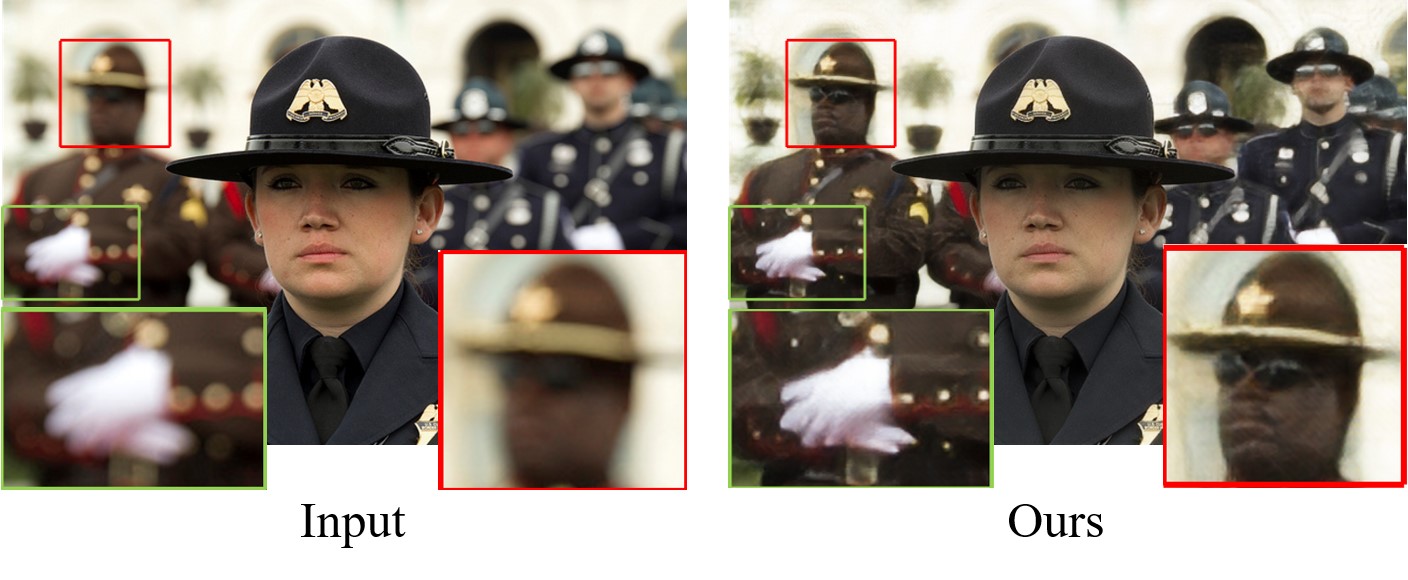} % Reduce the figure size so that it is slightly narrower than the column. Don't use precise values for figure width. This setup will avoid overfull boxes.
\caption{Defocus deblurring results of the proposed method on the CUHK dataset.}
\label{fig8}
\end{figure}

As can be seen in Table 4, the DG part is very important for defocusing the single-image, providing defocusing clues, and greatly improving the performance of the model, but the removal effect of large-area blur is still insufficient, as shown in Figure 5. Therefore, we introduce the MS module, which improves the performance of large-area defocus deblurring. Using L1-Loss will cause the image to be too smooth. For the deblurring task, it is important to restore the details and texture of the image. We introduce GAN and perceptual loss to further restore the structure and texture of the image, making the image closer to human perception.

\textbf{Effectiveness of learnable blur kernel.} 
To demonstrate the effectiveness of the proposed learnable BK, we analyzed the effects of different BKs on the recovered defocus map, as shown in Table 5. It can be seen that the defocus map generated using the learnable BK gets better performance. Since \cite{Abuolaim2021RDPD} does not provide specific parameters for sampling blur kernels, we can only perform qualitative comparisons. As can be seen from Figure 6, the real blur kernel approximates a band-stop filter, and we have learned this feature well.

\textbf{The impact of different GANs.} 
We also experiment with different GANs, as shown in Table 6. We use the adversarial loss without defocus weights as the original loss. Each pixel has the same weight, with indiscriminate attention to images. For example, using FPN \cite{lin2017FPN} in deblurGANV2 as G, the performance of PSNR 24.23 and LPIPS 0.150 is obtained. This shows that for the defocus deblurring task, to design a network for its characteristics, it is especially necessary to provide defocus clues. We also experiment with doubleGAN \cite{kupyn2019deblurGAN}, D that fuses global and local, but it does not work well. We think that the defocus blur is regional, not global, so D with global properties does not work well. While our proposed defocus adversarial loss increases the focus on defocused areas and achieves better performance.

\subsection{Generalization ability}
Generalizability is very important for a model. After training on the DPDD dataset, we test it directly on the Google PixelDP dataset and the CUHK dataset. Since there is no GT, we only show the qualitative results of the model. Figure 7 shows the results of the CUHK dataset. The results on the Google PixelDP dataset can be viewed in the supplementary materials. As we can see, our perception is very good compared to the input. For the CUHK dataset, which mainly contains portraits, we can significantly recover details and textures, such as the soldier's glasses, face, and hand contours.

\section{Conclusion}
We propose a single-image defocus deblurring GAN and an unsupervised method for estimating defocus maps with a learnable blur kernel. The learned defocus map is used to guide the network for defocus deblurring. The proposed network can effectively handle large-area blur and effectively reconstruct image details and textures. The recovered defocus maps are comparable to current supervised methods. The effect of each component is verified experimentally, and it accompanies a great performance with fewer parameters. 

With the development of deep learning, the introduction of mask learning and diffusion models has promoted progress in the field of image generation. Finally, utilizing a new framework to facilitate defocus deblurring research as future work would be of great interest.

\section*{Acknowledgements}
This work is supported by the Shenzhen Science and Technology Research and Development Fund (No. JCYJ20180503182133411, JSGG202011022153800002, KQTD20200820113105004). Jucai Zhai would like to thank Dr. Emad Iranmanesh for proofreading the article and for his constructive feedback.

\bibliography{aaai23.bib}

\end{document}